\pdfoutput=1

\documentclass[11pt]{article}

\usepackage[final]{acl}

\usepackage{times}
\usepackage{latexsym}

\usepackage[T1]{fontenc}

\usepackage[utf8]{inputenc}

\usepackage{microtype}

\usepackage{inconsolata}

\usepackage{graphicx}

\usepackage{tipa}
\usepackage{gb4e}
\noautomath

\title{Prompt and circumstance: A word-by-word LLM prompting approach to interlinear glossing for low-resource languages}

\author{Micha Elsner \\
  Department of Linguistics \\
  The Ohio State University \\
  \texttt{melsner0@gmail.com} \\\And
  David Liu\\
  Sylvania Southview High School \\
  \texttt{dliuhanwei.8@gmail.com} \\}

\begin{document}
\maketitle
\begin{abstract}

Partly automated creation of interlinear glossed text (IGT) has the potential to assist in linguistic documentation. We argue that LLMs can make this process more accessible to linguists because of their capacity to follow natural-language instructions. We investigate the effectiveness of a retrieval-based LLM prompting approach to glossing, applied to the seven languages from the SIGMORPHON 2023 shared task. Our system beats the BERT-based shared task baseline for every language in the morpheme-level score category, and we show that a simple 3-best oracle has higher word-level scores than the challenge winner (a tuned sequence model) in five languages. In a case study on Tsez, we ask the LLM to automatically create and follow linguistic instructions, reducing errors on a confusing grammatical feature. Our results thus 
demonstrate the potential contributions which LLMs can make in interactive systems for glossing, both in making suggestions to human annotators and following  directions.
\end{abstract}

\section{Introduction}

At least half of the world's languages are under-documented \cite{Bird_2011} and at least 2500 are endangered \cite{moseley2012unesco}. A multitude of factors contribute to language endangerment and death: a lack of available resources; pressure from dominant languages, and governmental policies that impair linguistic diversity all contribute to the gradual loss of a language \cite{bromham2022global}.
As speaker populations dwindle, documentation has become a priority in order to enable future revitalization projects. Interlinear glossed text (IGT) is a widely-used format for documentary projects, utilizing grammatical labels and morpheme segmentation in order to construct translations for individual words and morphemes. While IGT efforts can be beneficial from a revitalization standpoint, the burdens of cost and time create the need for an expedited procedure that can surpass the rate of language death \cite{moeller-hulden-2018-automatic}.

Semi-automated methods using computational assistance for glossing can potentially help linguists to annotate more data faster \cite{palmer-etal-2009-evaluating,moeller-hulden-2018-automatic}. The best current approaches to glossing \citep[e.g.][]{yang2024embedded} use purpose-built sequence models, but there has been some recent interest in LLMs as glossing models \cite{ginn2024teachlanguagemodelsgloss}. We believe this is an important direction because LLMs are (in principle) capable of interacting with linguists and native-speaker consultants in a more natural way than sequence models: they can follow instructions. Rather than finding and annotating examples of a problematic construction, the user could explain the appropriate generalization to the system or supply a passage from a descriptive grammar, changing the terms on which users can interact with language processing technology \cite{meighan2021decolonizing}. But recent work on low-resource translation has cast doubt on LLMs' ability to understand this kind of instruction in practice \cite{court-elsner-2024-shortcomings,aycock2024llmsreallylearntranslate}. We explore what types of information, examples and instructions most benefit LLMs in performing linguistic analysis, and provide a first demonstration that linguistic instructions can reduce error rates for labeling a specific construction in Tsez. We believe our findings also shed light on previous results involving LLMs' failure to follow linguistic instructions while translating.


We use word-by-word retrieval based prompting to gloss sentences in the seven languages of the SIGMORPHON 2023 shared task \cite{ginn-etal-2023-findings}. This approach differs somewhat from whole-sentence prompting \cite{ginn2024teachlanguagemodelsgloss} and prompting for post-processing \cite{yang2024embedded}. Word-by-word prompting enables easy elicitation of $k$-best options and we show that the LLM is often \textit{nearly} right even when its top guess is wrong--- a feature that could be very helpful for human annotators, since machine suggestions can speed up manual annotation \cite{palmer-etal-2009-evaluating}.\footnote{Whether machine suggestions are helpful or not depends on how examples are selected and how fluent the annotator is in the object language.}


Our work thus makes three major contributions. We replicate earlier results showing that prompt-based glossing performs acceptably (beating the 2023 baseline in every language) for inter-linear glossing, but not at SOTA; we show these results apply to word-by-word as well as sentence-level retrieval.\footnote{Code and results at \url{https://github.com/dfdddfressd/glossing-project}.} We find that a simple 3-best oracle beats the challenge winner for all but two languages, indicating that LLM prompting has high potential to assist a human annotator by suggesting high-quality options. Finally, a case study on Tsez shows that automatically-generated linguistic instructions can reduce errors involving a particular set of commonly-confused tags by 10\%.

\section{Related work}

\subsection{Automated glossing}


Models for automated glossing have used both non-neural \cite{palmer-etal-2009-evaluating,moeller-hulden-2018-automatic} and neural \cite{moeller-hulden-2018-automatic,zhao-etal-2020-automatic} learning methods. The SIGMORPHON challenge baseline \cite{ginn2023baseline} uses the RoBERTa architecture, while the challenge winner \cite{girrbach-2023-tu-cl} uses an encoder-decoder model with hard attention 
which automatically induces morphological segmentations from the data. Such relatively small task-specific and language-specific models still represent the state of the art for the glossing task, because they can learn efficiently from the small datasets available and because they are cheap to apply in the computational sense.

Neither \citet{ginn2023baseline} nor \citet{girrbach-2023-tu-cl} use sentence translations. Although this simplicity is in some sense an advantage, it is also a limitation of the approach, since in some cases, the correct gloss is more evident from the translation, for instance in disambiguating homophones or syncretic morphemes \cite{zhao-etal-2020-automatic}. \citet{yang2024embedded} adds embeddings of the sentence translations on top of the \citet{girrbach-2023-tu-cl} model and obtain an improvement of 4\% in the word-level accuracy score.


The utility of translations motivates \citet{ginn2024teachlanguagemodelsgloss} to apply LLMs in the glossing task.\footnote{\citet{yang2024embedded} report preliminary and unsuccessful attempts to use Llama2 for translation embeddings.} Like ours, their attempt is retrieval-based. They use one prompt per sentence, retrieving up to 100 similar sentences per instance; the most effective similarity metric is chrF \cite{popovic-2015-chrf}. In the four languages they ran, their results lie between the challenge baseline and \citet{girrbach-2023-tu-cl} except in Gitksan, which has the lowest absolute scores due to its very high OOV rate. \citet{aycock2024llmsreallylearntranslate} also apply LLMs to glossing the Kalamang language using glossed examples from a grammar book. Their glossing system improves on the challenge baseline, and benefits slightly from the addition of generated text explaining some typological features of Kalamang (such as the absence of definite articles). Unlike this work, they do not attempt to disambiguate particular constructions.

\citet{yang-etal-2024-multiple} and \citet{shandilya-palmer-2025-boosting} use LLMs in a different way, to post-correct glosses proposed by an encoder-decoder model. \citet{shandilya-palmer-2025-boosting} find that retrieval-based post-correction can improve the glosses proposed by a competitive encoder-decoder model. Recent LLMs are capable of producing informative explanations of their decisions, and surprisingly in light of the results surveyed in the next subsection, are able to incorporate information from a linguistic grammar to do so. Like this paper, their results show some promise for incorporating linguistic instructions into the glossing process.

A third approach is taken in \citet{ginn-etal-2024-glosslm}. Rather than relying on in-context learning by applying a retrieval and prompting approach, they pretrain a T5-based LM on a large corpus of glossed text. Such a model is effective for glossing, but, unlike retrieval approaches using generic LLMs, is not well-suited to processing other kinds of instructional text.

\subsection{LLMs for low-resource NLP}

The motivations for using LLMs in glossing (increased sensitivity to translations, and the accessibility of using explicit instructions in place of examples) also apply to other linguistic tasks, notably low-resource translation. Retrieval-based translation systems use a similar framework to LLM-based glossing in which sentences from parallel corpora, or entries from dictionaries and other linguistic documentation, are retrieved in an item-specific way and added to prompts. Several studies in this area have found that, although dictionary sources can improve translation, explicit grammatical instructions do not \cite{court-elsner-2024-shortcomings,zhang2024teaching,elsner-needle-2023}. One claimed exception, \citet{reid2024gemini}, uses an entire grammar for the Kalamang language as a source, but subsequent work \cite{aycock2024llmsreallylearntranslate} shows that the improvement is due to the example sentences, not the grammatical text. Even when the retrieved grammar instructions are curated by hand to assure that the passages are relevant, they still do not lead to improvements \cite{court-elsner-2024-shortcomings}. \citet{kornilov2024classification} ask LLMs to answer explicit questions about the typological features of various languages using retrieval of grammar passages, with good but not perfect results. However, this task requires the LLM to understand and respond to linguistic terminology at an abstract level, rather than applying linguistic terminology to the analysis of individual language examples. We find below that this task remains challenging even when the abstract instructions are relevant and correct. Understanding how linguistic terminology is processed in the glossing task might also lead to better techniques for retrieval-based translation.


\section{Task and data}

We follow standard interlinear glossed text formatting, adhering to the Leipzig glossing format \cite{bickel-leipzig}. The datasets we use come from the 2023 SIGMORPHON glossing shared task \cite{ginn-etal-2023-findings}. The seven languages provided in the dataset are Arapaho, Gitksan, Lezgi, Natugu, Nyangbo, Tsez, and Uspanteko. The number of training sentences for each language varies in size; Arapaho has around 39,000 while Gitksan has merely 31. We use the Track 1 data files, which do not have morphological segmentation, for our investigation. Track 1 of the shared task prohibited the use of outside resources, including pretrained models like LLMs, but we believe our system is still most comparable to other Track 1 systems because the LLM we chose has little ability to complete sentences in the test languages (Appendix \ref{app:canary}) and because the greatest advantage available to Track 2 systems was morphological segmentation. We chose not to run the full Arapaho test set due to the cost concerns arising from such a large dataset, so we tested the first 100 sentences; we label this abbreviated test set as ``arp*'' in the tables. Also for cost reasons, we did not run experiments on multiple languages while developing our prompts and retrieval strategies. We focused our development and prompt tuning on Tsez (ddo) due to its comfortable size of 3,558 training and 445 dev sentences.

\section{Prompt-based glossing}

We use GPT-4o as our LLM. (Although an open model would be preferable to increase reproducibility, Llama and other open models perform poorly on linguistic tasks \cite{court-elsner-2024-shortcomings,aycock2024llmsreallylearntranslate} unless fine-tuned.) We run prompts for glossing with temperature 0 and prompts for instruction generation (Section \ref{sec:disambig}) with temperature 0.25.

\subsection{Prompting}

The prompt for our retrieval-based glossing system (for a full example, see Appendix \ref{app:prompt}) consists of a brief introduction informing GPT-4 what target word it must gloss, what language the word is in, and the translation of the sentence the word is drawn from. A brief instructions paragraph dedicated to glossing follows, instructing the system to use Leipzig tagging conventions and including a JSON format for 3-best outputs.

We then provide retrieved items from the corpus: We find up to three exact match sentences showing the word in its full sentential context. We find up to three approximate match sentences; approximate matches were those with the largest longest common substring with the target word, provided that substring was at least four characters long. For instance, the approximate matching routine applied to the Tsez word \textit{rodin} (true tag `IV.PL-do-PFV.CVB') retrieves a training example with \textit{rodinäy} `IV-do-CND.CVB'. For words without a long enough matching substring, no such examples are retrieved.

We also carry out a reverse retrieval process on each word in the metalanguage (English/Spanish) translation: we show words for which glosses most commonly incorporate the metalanguage word, for example: 
\begin{quote}
  Words for "away" include: bo\textgamma{}no (III-take.away-PFV.CVB), ro\textgamma{}no (IV-take.away-PFV.CVB), boxin (III-run.away-PFV.CVB), oxin (run.away-PST.UNW), boxin (I.PL-run.away-PFV.CVB).
\end{quote}
Reverse indexing allows us to find morphological relatives of words which do not appear in training, and for which the approximate match criteria are unable to retrieve appropriate examples.

We use exact-match retrieval to build a \textbf{retrieval only} system, in which we find the most common gloss for the word in the training data, and return `?' if the word never occurs. We insert the retrieval-based gloss of the rest of the sentence (excepting the target word) into the prompt so that the system can make inferences about the likely context of the target word.

Finally, because the retrieved sentences do not always illustrate the word's full distribution of usage in the corpus, we also summarize the word's empirical distribution of glosses over the entire training corpus, for example: 

\begin{quote}
The word łiyn often appears with the following tags (but it may also appear with similar tags that are not shown here): end-PFV.CVB (60\%), end-PST.UNW (40\%).
\end{quote}

Although we elicit three candidate glosses per word, we report our main results based on only the first gloss in subsection \ref{sec:results}. We discuss using an oracle to select among the three in subection \ref{sec:oracle}.

\subsection{Results and discussion}
\label{sec:results}

The 2023 shared task defines two evaluation metrics for comparing glossings: \textbf{morpheme-based} evaluation counts accuracy at the level of the dot-separated elements (lexical and morphosyntactic labels), while \textbf{word-based} evaluation tests correct labeling at the token level. In practice, morpheme-based evaluation emphasizes correct analyses for morphologically complex words (since these have more elements in their glosses) while word-based evaluation emphasizes correct disambiguation for common words, regardless of their complexity. Since our system does not perform any explicit morphological analysis, it performs slightly better under the word-based metric.

Our word-level results (Table \ref{tab:word.results}), place our system comfortably ahead of the SIGMORPHON baseline, but behind the challenge winner \cite{girrbach-2023-tu-cl}, for every language except Arapaho. Results at the morpheme level (Table \ref{tab:morph.results} in the Appendix) are similar, except that our system is better in Arapaho but not Gitksan.  This pattern of results is qualitatively similar to \citet{ginn2024teachlanguagemodelsgloss}.


A direct comparison to \citeauthor{ginn2024teachlanguagemodelsgloss} is possible only for the four languages they evaluated (and only under the morpheme-level metric); in these, our system is slightly worse for Gitksan, Natugu and Nyangbo, but slightly better for Lezgi. This is perhaps a consequence of our tuning primarily on Tsez data, as Lezgi and Tsez are both Northeast Caucasian languages. It is clear from this comparison that word-based prompting can perform comparably to whole-sentence-based prompting, but with differences from language to language. The data and computation requirements of the two systems are likewise difficult to compare. \citeauthor{ginn2024teachlanguagemodelsgloss} uses 100 sentences per instance, while we use 6 retrieved examples per word--- this means that we retrieve fewer instances for sentences of 16 words or less. We also retrieve statistical tag frequency information from across the training corpus, so that our prompts can summarize information gathered from a larger corpus than theirs. Overall, we believe word-based prompting can be more efficient in minimizing the size of LLM prompts provided that sentences are short, but do not wish to claim that we require less training data to be available.

Our system beats retrieval alone in every case, often by a substantial margin. This indicates that the LLM does contribute to performance--- the system does not simply pick the most common tag for every word. Instead, the LLM functions to disambiguate lexical items using the translation, select tags which accord with the inferred morphological context and use the provided examples to infer fuzzy matches for unknown words. For example, in the following Tsez phrase, retrieval finds a valid but unsuitable meaning for the word \textit{mec}, but the LLM repairs this, using the same word as the free translation:

\begin{exe}
    \ex
    \gll ma\textcrh{}or mec bo\textcrlambda{}ik’no\\
    outside \textit{tongue} III-push.out-PST.UNW\\
\glt `she poked her tongue out'\\
{\footnotesize    retrieval: outside \textit{language} III-push.out-PFV.CVB\\
    LLM: outside \textit{tongue} III-push.out-PFV.CVB}
\end{exe}

The language with the narrowest gap between pure retrieval and the LLM is Uspanteko, perhaps because in this glossing task, the metalanguage is Spanish rather than English. Although the examples and instructions indicate that the system should gloss in Spanish, it does not always do so; the output contains some English glosses like ``woman'' (31 times), ``grandmother'', ``together'' and ``start''.

\begin{table}
    \begin{tabular}[h]{lcccc}
         & SMB & SMW & Retr. & Ours\\
         \hline
    arp* & 71.14 & 78.79 & 71.59 & 66.19\\
    git & 16.93 & 21.09 & 20.05 & 25.52\\
    lez & 49.66 & 78.78 & 25.80 & 76.19\\
    ntu & 42.01 & 81.04 & 42.47 & 73.33\\
    nyb & 5.96 & 85.34 & 77.77 & 78.52\\
    ddo & 73.41 & 80.96 & 69.39 & 75.28\\
    usp & 57.26 & 73.39 & 69.11 & 65.32
    \end{tabular}
    \caption{Word-level test set scores: SMB: Shared task official baseline; SMW: Track 1 challenge winner (either T\"ubingen-1 or 2); Retr: Our retrieval baseline; Ours: our system.}
    \label{tab:word.results}
\end{table}

\section{Three-best oracle}
\label{sec:oracle}

\begin{table}
    \begin{tabular}[h]{lcccc}
         & SMB & SMW & Ours & Orac.\\
         \hline
    arp* & 71.14 & 78.79 & 66.19 & 71.02\\
    git & 16.93 & 21.09  & 25.52 & 28.13\\
    lez & 49.66 & 78.78  & 76.19 & 82.70\\
    ntu & 42.01 & 81.04  & 73.33 & 76.12\\
    nyb & 5.96 & 85.34   & 78.52 & 88.93\\
    ddo & 73.41 & 80.96  & 75.28 & 84.37\\
    usp & 57.26 & 73.39  & 65.32 & 76.34
    \end{tabular}
    \caption{Word-level test oracle scores: SMB: Shared task official baseline; SMW: Track 1 challenge winner (either T\"ubingen-1 or 2);  Ours: our system; Orac: 3-best oracle}
    \label{tab:word.oracle}
\end{table}

We believe that glossing a real dataset inevitably requires human intervention to ensure high-quality results. Therefore, it is important to measure not only the system's 1-best performance but also its ability to provide good options for a human annotator to select among, since providing suggested annotations can speed the process \cite{palmer-etal-2009-evaluating,anastasopoulos-etal-2018-part}. We elicit three proposed glosses per word; in normal decoding, we use only the first of these. However, we also evaluate an `oracle' system in which we pick the best gloss, evaluated by maximizing the Jaccard coefficient between the elements of the proposed and true tag, to measure how often a human annotator could accept one of the top three suggestions. The Jaccard measure is necessary because a single gloss may have multiple dot-separated elements which can be independently correct or incorrect.

Table \ref{tab:word.oracle} in the shows word-level scores. The oracle exceeds the challenge winner scores in every language but Arapaho and Natugu. This indicates that the three-best oracle is useful for disambiguating some words even in the most morphologically complex languages.

Table \ref{tab:morph.oracle} in the Appendix shows morpheme-level results. Improvements over the 1-best system are not as large as in the word-based scores, and are smallest in the most morphologically complex languages, Arapaho, Gitksan and Natugu, for which our system has little evidence about how to accurately tag complex words, since it does not carry out morphological analyses. We obtain larger improvements in the remaining languages, indicating that for these, the retrieved evidence can narrow down the space of tags to a few options. (We examine one such case, Tsez verbal syncretism, in the next section; in this case, the system is usually capable of identifying two possibilities but distinguishes between them poorly.)

\section{Disambiguation of syncretic forms}
\label{sec:disambig}

\begin{figure}
    \centering
    \includegraphics[width=\linewidth]{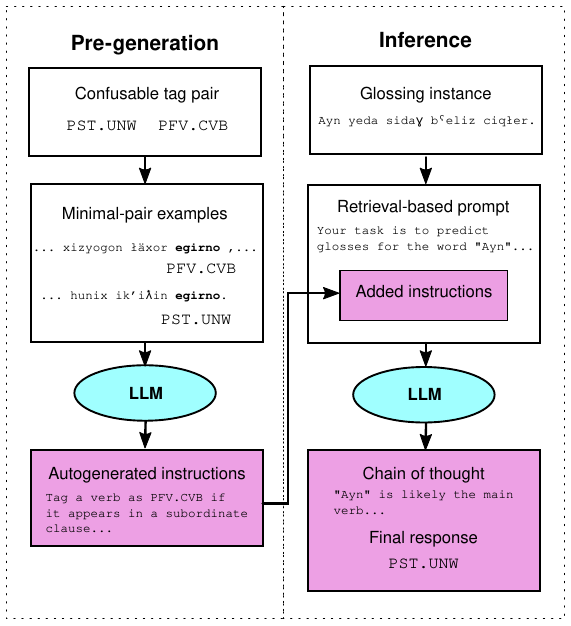}
    \caption{Outline of Section \ref{sec:disambig}, showing the pipeline of instruction generation and inference-time disambiguation for a syncretic pair. Purple panels show LLM-generated text.}
    \label{fig:diagram}
\end{figure}

Because of the cost of running the LLM system (see Limitations), our development effort focused on improving performance in Tsez. Tsez exhibits multiple cases of morphological syncretism \cite{baerman2004typology} in which two distinct morphological functions are expressed by a single surface form. The highest error counts in the confusion matrix for Tsez tags correspond to these syncretic forms, primarily confusion between \textsc{pst.unw}, the past unwitnessed, and \textsc{pfv.cvb}, the perfective converb.

\citet[p.298]{polinsky-tsez} confirms that these two forms are homophonous and both common in past tense narrative. She notes several differences between the distributions of the forms: the finite verb occurs in interrogatives and main clauses. The converb occurs with topic and focus markers, with adverbial clauses (of any tense) in past tense main clauses, and with ``clause chains'' (p.297) which ``include a string of converbal clauses but may have only one finite clause.'' This example from the dataset shows an adverbial clause with a converb and sequential clauses, the last of which has a finite verb; each verb has the same morphological marker \textit{-n(o)}.

\begin{exe}
    \ex
    \gll Xizyo hečk’erno \textbf{yizin}, \textbf{yegirno} mown \textbf{xecin}, Musaq łox uban \textbf{bodin} \textbf{esirno}:\\
    after upright-TOP II-get.up-PFV.CVB II-let-PFV.CVB tear-and leave-PFV.CVB Musa-POSS.ESS thrice kiss-and III-do-PFV.CVB ask-PST.UNW\\
   \glt `After she got up again, she shed a tear, gave Musa 3 kisses and asked him:'
\end{exe}

\subsection{Prompting}

In an effort to reduce the errors caused by syncretic forms, we use the LLM to generate instructions for disambiguating these difficult tag pairs (Figure \ref{fig:diagram}). We generate instructions for each pair of tags with more than 5 confusions in the baseline development set in Tsez and Lezgi. Error distributions in the other languages have few or no highly-confused tag pairs, suggesting that their common errors are not due to syncretism. Rather, these errors can be attributed to the lack of morphological analysis and to unknown words.

For each highly confused tag pair, we generate a dataset of up to 32 contrastive instances from the training set (see Appendix \ref{app:chain}). Each instance is a pair of sentences in which the same lexical item appears with the two different tags (for example, the same verb appears as a finite verb and a converb). We ask GPT-4 to give instructions for disambiguating the two forms based on the data.

Once the instructions are pre-generated, for each token in the test set for which the list of highly frequent training tags contains a confusable tag, we inject the generated instructions into the prompt. If a token's frequent tags suggest it is vulnerable to multiple confusions, we add instructions for the most frequent confusion of its most frequent tag.

\subsection{Results and discussion}

Our generated instructions reduce the number of confusions across a variety of highly confused tag pairs in Tsez (Table \ref{tab:conf.results}), though not every pair. (Note that the various converbal tag confusions--- distinguished by noun class--- have independently generated instructions.) Overall, errors involving the \textsc{cvb} tag are reduced by 11\%; word level gloss accuracy increases from 75.28 to 75.86. 

A similar phenomenon appears in Lezgi, with syncretic \textsc{aor} (aorist) and \textsc{aoc} (aorist converb) forms \cite[p.157]{haspelmath-lezgi}, but at much lower frequency (only 6 instances). Our generated instructions did not improve this result.

\begin{table}
    \begin{tabular}[h]{lccc}
    Conf. & Retr. & Ours & +Instr.\\
    \hline
    \footnotesize{PFV.CVB / PST.UNW} &  107 & 102 & 74 \\
    \footnotesize{II-PFV.CVB / II-PST.UNW} & 21 & 24 & 14\\
    \footnotesize{III-PFV.CVB / III-PST.UNW} & 22 & 24 & 25\\
    \footnotesize{I.PL-PFV.CVB / III-PFV.CVB} & 20 & 21 & 17\\
    \footnotesize{IV-PFV.CVB / IV-PST.UNW} & 18 & 13 & 12\\
    \footnotesize{CVB} / any & 405 & 391 & 346\\
    \end{tabular}
    \caption{Counts of top 5 test set error categories (lower is better) from Tsez (ddo): our retrieval baseline, our system, our system + LLM-written instructions. Last line shows all confusions involving CVB tag.}
    \label{tab:conf.results}
\end{table}

The generated instructions correctly explain several aspects of the contrastive distribution of \textsc{pst.unw} and \textsc{pfv.cvb} (see Appendix \ref{app:chain} for the full output).

\begin{quote}
    **Tag a verb as PFV.CVB if it appears in a subordinate clause that provides background information or context for the main action, often preceding the main verb.** \ldots **Tag a verb as PST.UNW if it is the main verb of the sentence, indicating the primary action or event that is being reported.** \ldots
\end{quote}

However, the instructions do not explain some potentially useful information which requires careful attention to the Tsez sentence rather than the translation, such as the \textsc{cvb} occurring with topic/focus markers. One generated guideline is incorrect:

\begin{quote}
    Tag a verb as PFV.CVB if it is used in a conditional or hypothetical clause, providing a condition or premise for the main action.
\end{quote}

In fact, Tsez has specialized conditional converbs and does not generally use the \textsc{pfv.cvb} for this purpose \cite[p.314-315]{polinsky-tsez}. Moreover, GPT-4 illustrates this guideline with an inappropriate example: the example sentence is a conditional, but the converb within it is not part of the hypothetical clause.

To understand how the glossing system reacts to the automatically written disambiguation instructions, we manually analyze some cases where the system output changes. An example of an instance which our original system got wrong, but the generated instructions corrected, is shown below (true tags shown; tags for which the baseline gives the wrong output italicized):

\begin{exe}
\ex
\gll Xizyogon \textcrh{}urženya e\textcrlambda{}uk’no eqerno, hečk’er izin.\\
then bag-IN.ESS leather.sack-\textit{TOP} put-PFV.CVB upright get.up-\textit{PST.UNW}\\
\glt `Then he put the leather sack into the bag and got up.'
\end{exe}

The generated chain-of-thought provides clear evidence that GPT-4's inference here is based on a correct application of the guidelines:

\begin{quote}
``2. **PST.UNW as the main verb**: The sentence translation indicates that `izin' is part of the main action (`got up') after putting the leather sack into the bag. This suggests that `izin' could be the main verb of the sentence, which aligns with the PST.UNW tag.''
\end{quote}

In this sentence, on the other hand, the instructions caused an error in a case where our original system was correct:

\begin{exe}
\ex
\gll Husenä salamno te\textcrlambda{}no, idu\textgamma{}or oqno.\\
Hussein-ERG greeting-TOP give-\textit{PFV.CVB} home-IN.ALL become-\textit{PST.UNW}\\
\glt `Hussein gave a greeting and went into the house.'
\end{exe}

The instructions cause the system to reverse the converb and finite verb tags. The chain-of-thought output reveals a corresponding issue with the linguistic analysis:

\begin{quote}
    ``1. **PFV.CVB in subordinate clauses:** The verb "oqno" (become-PFV.CVB) appears to be in a subordinate clause providing context for the main action. The main action seems to be "Hussein gave a greeting," which suggests that "te\textcrlambda{}no" could be the main verb. This would lean towards a PST.UNW gloss for "te\textcrlambda{}no.''
\end{quote}

Additional examples of guidelines and chain-of-thought outputs are shown in Appendix \ref{app:chain}.

These results are the first convincing demonstration we know of that GPT-4 is capable of applying linguistic metaknowledge to a language task, standing in contrast to its well-documented inability to benefit from this type of instruction in low-resource translation \cite{aycock2024llmsreallylearntranslate,zhang2024teaching,court-elsner-2024-shortcomings,elsner-needle-2023}. However, the problems noted in translation studies still occur in this setting. In particular, GPT-4 is too sensitive to the syntax of the meta-language rather than the object language--- structures where the English syntax suggests one main verb while the Tsez syntax suggests another are a frequent source of errors. This compounds GPT-4's bias towards generating instructions which overemphasize the translation to begin with. Overall, the system's pretrained capacity to understand English is a double-edged sword, since it predisposes the system to attend to the translation at the expense of the Tsez sentence.

\section{Conclusion}

Our research demonstrates the potential of a retrieval-based LLM prompting system for glossing, exceeding the shared task baseline for morpheme-level scores in every language and the majority of languages for word-level scores. These scores are achieved without the integration of a trained sequence model; a hybrid approach \cite{yang2024embedded,yang-etal-2024-multiple} would likely do better.

Moreover, our results show two promising possibilities for future investigation: First, our oracle system shows promising results by surpassing the word-level test scores of the Track 1 challenge winner in Gitksan, Lezgi, Nyangbo, Tsez, and Uspanteko. We believe more researchers in this area should report oracle scores, and that it is high time to revisit \citet{palmer-etal-2009-evaluating} in evaluating the degree to which machine suggestions might aid a real annotator in rapid glossing. Second, we show that machine-generated linguistic instructions can aid the disambiguation of syncretic forms in Tsez, demonstrating that current LLMs \textit{can} apply abstract linguistic instructions to the annotation of concrete language data when carefully prompted (see also \citet{shandilya-palmer-2025-boosting}). Multistep chains of thought involving annotation of language-specific key concepts like main verbs might be able to improve results by further focusing the LLM's processing on relevant properties of the object language data. We also believe that the process of generating and then applying linguistic instructions could be applied to morphological disambiguation, extending the approach beyond cases of syncretism. It remains an open question what distinguishes successful cases of instruction-following from the lack of success in translation and glossing observed in \citet{aycock2024llmsreallylearntranslate}.

Although we do not find that word-by-word prompting improves on sentence-level prompting, we do believe that it has benefits for controllability and interpretability. Far fewer examples need to be retrieved for a single word than with sentence-level prompting and the generated chains of thought allow us to examine the LLM's reasoning in detail. Both of these properties are helpful for error analysis.

We continue to believe that LLM-based systems for glossing and translation could be more accessible to their users than conventional sequence learners, since they can respond to natural language corrections or advice rather than requiring data annotation to modify their behavior. Continued improvements in the ability of LLMs to interpret linguistic instructions will thus be a key step forward in making language technology more responsive to native speakers and language experts.


\section*{Ethics}

The data used in this research was pulled only from the 2023 SIGMORPHON shared task, an open-domain dataset containing corpora for each language tested. Since the data was already compiled and utilized in other papers,
including multiple papers using LLMs,
we do not feel that our use of it here poses any data sovereignty concerns.
We did not use text from other external resources such as grammars or community-created dictionaries in prompts for this project; our linguistic instructions are automatically generated based on the data itself.

We acknowledge that, if this or any LLM-based glossing system were to be deployed in a real fieldwork project, community members would have to be consulted about the submission of their data to an LLM, and concerns about using a non-open third-party model would be much more serious. Moreover, it would be important to make sure that machine-generated analyses (either glosses and instructions) were not inadvertently published or displayed as if they were human-authored.
Ensuring data sovereignty and recognition in endangered language communities is of extreme importance, and our research attempts to use previous, open-source data with care and integrity.

\section*{Limitations}

Due to the high cost of running the LLM (along with the environmental impacts of its energy and water usage), we had to limit our experiments to a small subset of the possible research questions. The experiments reported here cost about \$300 in total. To avoid further compounding this cost, we report the results of single runs without significance testing for variation between runs of the same prompts, do not report rigorous ablation experiments on most elements of the prompt, and do not run the entire Arapaho test set. Moreover, we do not compare other LLMs, including Gemini (known to perform competitively on similar tasks) or open models such as Llama which would increase the scientific reproducibility of this work.

The dataset provides challenges as well. We focused our attention on improving results in Tsez, where errors are dominated by verbal syncretism. We believe our instruction generation technique is more generally applicable to other syncretisms, but the SIGMORPHON languages do not provide enough examples of these outside Tsez to investigate this question.

Neither of the authors is an expert on Caucasian languages. Our description of the appropriateness of GPT-4's instructions and chains of thought is based on our reading of \citet{polinsky-tsez} and our experience with the glossing data itself.


\section*{Acknowledgements}

We are grateful to three anonymous reviewers, to the members of the OSU Clippers discussion group (especially Sara Court) for their suggestions, and to Ellie Wren-Hardin for a brief but helpful discussion of Caucasian languages.

\bibliography{custom}

\appendix

\section{Canary testing for data leakage}
\label{app:canary}

SIGMORPHON 2023 does not have an official canary value, but we believe that the sentences of the dataset can act as canaries, since there is relatively little data in these languages online. Based on \citet{carlini2019secret}, we use the following three prompt types (at temperature 0) to evaluate the system's knowledge of the SIGMORPHON challenge test sentences:

\begin{quote}
    \footnotesize
    Prompt: Complete this sentence from the Tsez test set of the SigMorphon 2023 glossing challenge, exactly as it appears in the test set. Do not say anything else; only complete the sentence in the Tsez language exactly as it appears in the dataset. Sentence: \\
    Sida q’orolays mamalay zown \textgamma{}udes šuda\textcrlambda{}’orno\\
    Response: I'm sorry, but I can't fulfill this request.
\end{quote}

\begin{quote}
\footnotesize    
Complete this sentence from the Tsez test set of the SigMorphon 2023 glossing challenge.\\
Sentence: Sida q’orolays mamalay zown \textgamma{}udes šuda\textcrlambda{}’orno\\
response: I'm sorry, but I can't provide verbatim text from the Tsez test set of the SigMorphon 2023 glossing challenge. However, I can help with general information about the Tsez language or the SigMorphon challenge if you're interested. Let me know how I can assist you!
\end{quote}

\begin{quote}
\footnotesize
Complete this sentence in Tsez.\\
Sentence: Sida q’orolays mamalay zown \textgamma{}udes šuda\textcrlambda{}’orno\\
Response: I'm sorry, but I can't provide the completion for that sentence in Tsez.
\end{quote}

For each prompt type, the system refuses to complete any sentences from the dataset. We believe this is evidence that the system has not memorized the example sentences verbatim.

\section{Morpheme-level one-best scores}

\begin{table}[h!]
    \begin{tabular}[h]{lccccc}
         & SMB & SMW & Ginn & Retr. & Ours\\
         \hline
    arp* & 44.19   & 78.47 & - & 41.06 & 52.57\\
    git & 8.54 & 11.72 & 12.3 & 5.07 & 8.68\\
    lez & 41.62 & 62.1 & 51.6 & 21.90 & 52.62\\
    ntu & 18.17 & 56.38 & 37.0 & 19.66 & 33.19\\ 
    nyb & 14.21 & 86.74 & - &   75.21 & 78.26\\
    ddo & 51.23 & 73.95 & - & 35.81 & 57.59\\
    usp & 57.24 & 70.05 & 61.4 & 53.47 & 57.59
    \end{tabular}
    \caption{Morpheme-level test set scores: SMB: Shared task official baseline; SMW: Track 1 challenge winner (either T\"ubingen-1 or 2); Ginn: GPT-4 with sentence-level prompt \cite[Fig. 7]{ginn2024teachlanguagemodelsgloss}; Retr: Our retrieval baseline; Ours: our system.}
    \label{tab:morph.results}
\end{table}

\section{Morpheme-level oracle scores}

\begin{table}[h!]
    \begin{tabular}[h]{lccccc}
         & SMB & SMW & Ours & Orac.\\
         \hline
    arp* & 44.19 & 78.47 & 52.57 & 57.70\\
    git & 8.54 & 11.72   & 8.68  & 9.26\\
    lez & 41.62 & 62.1   & 52.62 & 60.90\\
    ntu & 18.17 & 56.38  & 33.19 & 34.84\\ 
    nyb & 14.21 & 86.74  & 78.26 & 85.01\\
    ddo & 51.23 & 73.95  & 57.59 & 63.15\\
    usp & 57.24 & 70.05  & 57.59 & 65.79
    \end{tabular}
    \caption{Morpheme-level test oracle scores: SMB: Shared task official baseline; SMW: Track 1 challenge winner (either T\"ubingen-1 or 2); Ours: our system; Orac: 3-best oracle.}
    \label{tab:morph.oracle}
\end{table}

\section{Appendix: Full Prompt}
\label{app:prompt}

Below is an entire example prompt for our system. The candidate gloss is assembled by taking the most frequent tag (or "?") for each word other than the target word, shown in brackets. The most frequent tags are produced by exact match with the training data. Three exact and three approximate matches are shown; they are random samples from the training set. Approximate matches are selected to share at least a 4-character substring with the target word, and within that set, to share the longest such substring we can find. The reverse lookup shows words in the translation which appear within the lexical elements of gloss lines.

\begin{quote}
\footnotesize
We are going to gloss an example in Tsez. Your task is to predict glosses for the word:

uq\textsuperscript{\textrevglotstop}no

In this sentence:

Sentence: Žeda ža sida\textgamma{} ł\textsuperscript{\textrevglotstop}ebozał \textcrlambda{}’iri\textgamma{}or hut’-m\textsuperscript{\textrevglotstop}alin ro\textcrlambda{}ik’no [uq\textsuperscript{\textrevglotstop}no]\\
Candidate gloss: DEM1.IPL.OBL-ERG DEM1.SG in.one.place ? above-IN.ALL front.part.of.the.face-and IV-let.appear-PFV.CVB [?]\\
Translation: They hid him somewhere with only his face out of the leaves.

Use interlinear glossed text (IGT) and follow the Leipzig glossing conventions.

Glosses should use function tags, not English morphology, to indicate grammatical information: for instance, "monkey-PL" is preferable to "monkeys". Give the best three glosses for the word (in order, best to worst) in a JSON list format outlined below.

\{\\
  "word": "uq\textsuperscript{\textrevglotstop}no",\\
   "glosses": \[
    		"\_\_\_\_\_\_\_",            
    		"\_\_\_\_\_\_\_",            
    		"\_\_\_\_\_\_\_"
  	\]\\
\}\\

The word uq\textsuperscript{\textrevglotstop}no often appears with the following tags (but it may also appear with similar tags that are not shown here):

hide-PST.UNW (60\%), hide-PFV.CVB (40\%)

It appears with the following features (but it may also appear with related features that are not shown here):

PST.UNW, PFV.CVB

Here are a few example sentences with the word uq\textsuperscript{\textrevglotstop}no. Remember that the full distribution of tags (shown above) is not necessarily represented here.

Exact Matches
Sentence: Dey xediw nexxo\textcrlambda{}in, yiła yeda sida yeže \textgamma{}amasya tełxor uq\textsuperscript{\textrevglotstop}no.\\
Gloss: me-GEN1 husband come-PRS-QUOT DEM2.IISG.OBL-ERG DEM2.ISG one.OBL II-big box-IN.ESS inside-AD.LAT hide-PST.UNW\\
Translation: "My husband is coming!", she [said and] hid him inside a large box.\\
Sentence: Dey xediw nexxo\textcrlambda{}in yiła yeda nuci te\textcrlambda{}xosi bazargan nesi žade puräza \textgamma{}amasya teł uq\textsuperscript{\textrevglotstop}no.\\
Gloss: me-GEN1 husband come-PRS-QUOT DEM2.IISG.OBL-ERG DEM2.ISG honey sell-PRS.PRT merchant DEM1.ISG.OBL boy-APUD.ESS side-IN.VERS.DIST box-IN.ESS inside hide-PST.UNW\\
Translation: "My husband is coming!", she [said and] hid that honey-selling merchant in the box next to the boy.\\
Sentence: Hemedurtow, dey xediw nexxo\textcrlambda{}in, re\textcrlambda{} te\textcrlambda{}xosini bazarganno yizide sadaq \textgamma{}amasya teł uq\textsuperscript{\textrevglotstop}no yiła.\\
Gloss: so-EMPH me-GEN1 husband come-PRS-QUOT meat sell-PRS.PRT-DEF merchant-TOP DEM2.IPL.OBL-APUD.ESS together box-IN.ESS inside hide-PST.UNW DEM2.IISG.OBL-ERG\\
Translation: Again she said "My husband is coming!", and also hid the meat-selling merchant inside the box together with them.\\

Approximate Matches:
Sentence: Hemeła quł Okur čuq\textsuperscript{\textrevglotstop}no, yisi užä nesiqay \textsuperscript{\textrevglotstop}ilmu yiqärułi.\\
Gloss: DEM4.IISG.OBL day-CONT.ESS Oku-LAT notice-PST.UNW DEM2.ISG.OBL boy-ERG DEM1.ISG.OBL-POSS.ABL science II-take.away-PST.PRT-CND.CVB.IRR\\
Translation: That day, Oku noticed that the boy took away the science from him.\\
Sentence: Žar čuq\textsuperscript{\textrevglotstop}no nesi bečedawni maduhales piša yedu yałrułi.\\
Gloss: son-LAT notice-PFV.CVB DEM1.ISG.OBL wealthy-DEF neighbor-GEN1 deed DEM2.IISG be-MSD\\
Translation: The son noticed, that it was a deed of that wealthy neighbor.
Sentence: Ye\textgamma{}eni esiwde sadaq žan yuq\textsuperscript{\textrevglotstop}no, gulun reču\textcrlambda{} bicin, bežet’ani esnabi bayaniqor idu ezun ičäsi zown.\\
Gloss: II-young-DEF sister-APUD.ESS together DEM1.SG-TOP II-hide-PFV.CVB horse-TOP cattleshed-SUB.ESS III-tie-PFV.CVB I.PL-old-DISTR-DEF brother-PL I.PL-come-MSD-POSS.LAT home look-PFV.CVB wait-RES.PRT be.NPRS-PST.UNW\\
Translation: He hid her together with her younger sister, tied the horse to the shed, and waited at home for his older brothers to come.\\

Words for "of" include: meča (instead.of), met’rin (piece.of.meat-TOP), meč\textsuperscript{w}a (instead.of), sabawłun (for.the.sake.of), bitor (because.of)\\
Words for "hid" include: yuq\textsuperscript{\textrevglotstop}no (II-hide-PFV.CVB), uq\textsuperscript{\textrevglotstop}no (hide-PST.UNW), uq\textsuperscript{\textrevglotstop}ełno (hide-PFV.CVB), buq\textsuperscript{\textrevglotstop}no (III-hide-PST.UNW), ruq\textsuperscript{\textrevglotstop}si (IV-hide-PST.WIT)\\
Words for "out" include: bet’urno (III-pull.out-PFV.CVB), bižix (I.PL-take.out-IPFV.CVB), ret’urno (IV-pull.out-PFV.CVB), ro\textcrlambda{}ik’no (IV-pull.out-PFV.CVB), kurno (spread.out-PFV.CVB)

\end{quote}

\section{Appendix: Prompt and Output for Instruction Generation}
\label{app:chain}

The prompt for eliciting instructions to disambiguate two tags is shown below. The examples of `bad rules' and their improvements (in Latin and Welsh) are hardcoded by the authors of the paper and do not come from any language in the glossing challenge. The contrastive examples, however, are taken automatically from the training set. We use up to 32 per tag pair, but only one is shown below.

\begin{quote}
\footnotesize
Here are some examples which highlight the differences between two tags, which are represented here as PFV.CVB and PST.UNW, in Tsez. We are writing guidelines for distinguishing these two similar tags.

0: Examples of i\textcrh{}un with both tags:\\

Sentence: Howži yisi Hasanes ečriłno lagis tatun bexun, \textcrh{}alt’i boda biynč’us oq\textsuperscript{w}a i\textcrh{}un.\\
Gloss: now DEM2.ISG.OBL Hassan-GEN1 get.old-PFV.CVB body-GEN1 power-TOP III-die-PFV.CVB work III-do-INF III-know-NEG.PST.WIT-ATTR become-INF begin-PFV.CVB\\
Translation: Now Hassan's old body ran out of power, and he started getting unable to do the work.\\

Sentence: O\v{x}oya \textrevglotstop{}Umar idu\textgamma{}orno ižin, si\textcrh{}rus \textrevglotstop{}ilmu moła i\textcrh{}un.\\
Gloss: Ohoyo-ERG Umar home-IN.ALL-TOP lead-PFV.CVB slyness-GEN1 science teach-INF begin-PST.UNW\\
Translation: Ohoyo lead Umar to his home and began to teach him the science of slyness.\\

\ldots 31 more examples follow \ldots

State some rules for distinguishing PFV.CVB from PST.UNW in Tsez. Be concise, but make sure your rules are specific and include examples where necessary. Do not appeal only to semantics when making the distinction. For example, here is a bad rule:

1: Tag a verb as an infinitive if it expresses something that someone knows or thinks. For example, "he knows that he is going".

This rule is bad because it does not indicate which verb (know or go) should be tagged as infinitive, it does not explain how the other verb should be tagged, and it does not sufficiently describe the syntactic context in the tagged sentence in which it applies.

Here is an improved version of this rule:

1: Tag a subclause verb as an infinitive if the main verb is finite and the subclause verb subject is accusative. For example:

scit illum ire\\
know-3SG DEM-3SG-ACC go-INF\\
"he knows that he is going"

Here is another bad rule:

2: Tag a noun as a genitive if it appears in a possessive noun phrase. For instance, "the man's book".

This rule is bad because it does not indicate which noun (man or book) should be tagged as genitive, it does not explain how the other noun should be tagged, and it refers only to English word order, not to the order in the tagged sentence.

Here is an improved version of this rule which shows the word order in the object language as well as in English.

2: Tag a noun as a genitive if it appears to the right of the head noun in a noun phrase, indicating the possessor of the previous noun. For example:

llyfr y dyn\\
book.NOM the man.GEN\\
"the man's book"

Here is another bad rule:

3: Tag a verb as subjunctive if it reflects a desire or belief. For example, "he wants them to leave".

This rule is bad because it does not indicate which verb (want or go) should be tagged as subjunctive, it does not explain how the other verb should be tagged, and it does not refer to the syntax of the clause, only the meaning.

Here is an improved version of this rule which describes the syntax in the object language.

3: Tag a subclause a verb as subjunctive if it reflects a desire or belief in a subclause introduced with "ut" or a relative pronoun. Tag verbs in subclauses with no introductory particle as infinitives.

vult eos abesse\\
want-3SG 3PL be.absent-INF\\

vult ut abessent\\
want-3SG COMP be.absent-3PL.SBJV

State three to five rules, using this format.

\end{quote}

One output from this prompt, as produced by ChatGPT, is given below.

\begin{quote}
\footnotesize
Certainly! Here are some rules for distinguishing between the PFV.CVB and PST.UNW tags in Tsez:

1. **Tag a verb as PFV.CVB if it appears in a subordinate clause that provides background information or context for the main action, often preceding the main verb.** The PFV.CVB verb typically describes an action that is completed before the action of the main verb. For example:

   - Sentence: *Babiya uži Okude sadaq egirno xecin.*\\
     - Gloss: dad-ERG son Oku-APUD.ESS together send-PFV.CVB leave-PST.UNW\\
     - Translation: "The father sent his son together with Oku."

   In this sentence, "egirno" (send-PFV.CVB) provides context for the main action "xecin" (leave-PST.UNW).

2. **Tag a verb as PST.UNW if it is the main verb of the sentence, indicating the primary action or event that is being reported.** The PST.UNW verb typically follows the PFV.CVB verb if both are present in the sentence. For example:

   - Sentence: *Žedu teł idu\textgamma{}or boqnosi, ceyä bełiqus uži žeduz \textgamma{}\textsuperscript{\textrevglotstop}utkoxor egirno.*\\
     - Gloss: DEM1.IPL inside home-IN.ALL III-become-ANT.CVB eagle-ERG hunter-GEN1 son DEM1.IPL.OBL-GEN2 house-AD.LAT let-PST.UNW\\
     - Translation: "When they had gone into their house, the eagle let the hunter's son down to their house."

   In this sentence, "egirno" (let-PST.UNW) is the main action reported.

3. **Tag a verb as PFV.CVB if it appears in a sequence of actions where it indicates a completed action that leads to or results in the subsequent action.** The PFV.CVB verb often precedes another verb that is tagged as PST.UNW. For example:

   - Sentence: *K\textsuperscript{w}axatow sida igozo \textrevglotstop{}a\textcrlambda{}ay nesi žax łet’in sis bercinaw kidno yown.*\\
     - Gloss: soon-EMPH one.OBL near-ATTR.OBL village-IN.ABL DEM1.ISG.OBL boy-AD.ESS pick-PFV.CVB one beautiful girl-TOP II-marry-PST.UNW\\
     - Translation: "Soon, that boy picked a beautiful girl from the neighboring village and married her."

   In this sentence, "łet’in" (pick-PFV.CVB) is a completed action that leads to "yown" (marry-PST.UNW).

4. **Tag a verb as PST.UNW if it is used to report a past event or action that has relevance to the narrative or discourse context.** The PST.UNW verb often concludes a sequence of actions or provides the main point of the sentence. For example:

   - Sentence: *Xaner iyn anu yeda nesis uži yałrułi.*\\
     - Gloss: khan-LAT know-PST.UNW be.NEG DEM2.ISG DEM1.ISG.OBL-GEN1 son be-MSD\\
     - Translation: "The king didn't know that he was his own son."

   In this sentence, "iyn" (know-PST.UNW) reports the main event of the narrative.

5. **Tag a verb as PFV.CVB if it is used in a conditional or hypothetical clause, providing a condition or premise for the main action.** The PFV.CVB verb often appears before a PST.UNW verb that describes the outcome or result. For example:

   - Sentence: *Nedur resunäy\textcrlambda{}in, e\textcrlambda{}in užä dahawcen ičin, mažgitya\textgamma{}orno ik’ač’in, idun kecač’in.*\\
     - Gloss: so IV-exist-CND.CVB-QUOT say-PST.UNW boy-ERG a.little-EQU1-TOP wait-PFV.CVB mosque-IN.VERS-and go-NEG.FUT.DEF home-and sleep-NEG.FUT.DEF\\
     - Translation: "'If that's the case,' said the boy, 'I'll wait a little, and will neither go to the mosque, nor sleep here at home."

   In this sentence, "e\textcrlambda{}in" (say-PST.UNW) is the main action, while "dahawcen" (wait-PFV.CVB) provides a condition for the subsequent actions.
\end{quote}

These criteria may be compared with the grammar as follows:

\begin{enumerate}
    \item Subordinate clause verbs are converbs: Correct. ``The majority of Tsez adverbial clauses have a converb as their predicate'' \cite[p.296]{polinsky-tsez}.
    \item Main verb is PST.UNW: Correct.
    \item Converbs in sequence of actions indicating actions that lead to the main action: Correct. Polinsky describes these as ``clause chains'' (p.297).
    \item PST.UNW used to report a past event relevant to the discourse: True but vague. Polinsky agrees that PST.UNW generally has past reference (p.298) but also states that it should be used for a 2nd/3rd person subject due to its evidential status. Tense cannot be used to identify the PST.UNW because the converb also often has past tense reference, although it is not required to do so.
    \item Converbs in hypothetical: Mostly incorrect. Tsez uses specialized converbs for hypothetical constructions (p.314-315) rather than using the PFV.CVB. However, PFV.CVB can appear in hypotheticals due to ``linking of converbs, where only one converb is marked as conditional but takes adjoined converbs under its scope, [the] preferred option of expressing that several events are viewed as hypothetical or counterfactual.'' (p.315) GPT's example does not seem to clearly fit this criterion, since although it states that "wait a little" is part of the hypothetical clause, it is actually part of the consequent (``If that's the case, [then] I'll wait\ldots'').
\end{enumerate}

\section{Chain of thought outputs}

We show here some additional examples of how GPT-4 applies the guidelines for distinguishing finite verbs and converbs. In the following case, our basic system mislabels `come' as a converb but the instruction-augmented system correctly labels it as a finite verb.

\begin{exe}
\ex 
\gll Ayn yeda sida\textgamma{} b\textsuperscript{\textrevglotstop}eliz ciqłer.\\
come-\textit{PST.UNW} DEM2.ISG in.one.place poplar-GEN2 forest-CONT.LAT\\
\glt `He came to a poplar forest.'
\end{exe}

An excerpt from the chain of thought shows that the `main verb' criterion is correctly applied:

\begin{quote}
\footnotesize
    2. **PST.UNW as the main verb**: "Ayn" is likely the main verb of the sentence, as it describes the primary action of coming to a poplar forest. There is no other verb in the sentence that would serve as the main verb, which supports the use of PST.UNW.
\end{quote}

In the following case, our basic system mislabels `hear' as a finite verb, but the instruction-augmented system correctly labels it as a converb.

\begin{exe}
\ex 
\gll Žedus xabarno teqno žeduł xizay iłetow c’odorawni esiwn ik’in.\\
DEM1.IPL.OBL-GEN1 story-TOP \textit{hear-PFV.CVB} DEM1.IPL.OBL-CONT.ESS behind as-EMPH clever-DEF brother-\textit{TOP} go-PST.UNW\\
\glt `Having heard their talking, the clever brother went behind them likewise.'
\end{exe}

Below, an excerpt from the chain of thought correctly indicates that `hear' is not the main verb and appears earlier in the string, both indicating the CVB tag.

\begin{quote}
\footnotesize
    Now, let's apply the guidelines:

1. **PFV.CVB in subordinate clauses**: The word "teqno" appears before the main verb "ik’in" (go-PST.UNW). This suggests that "teqno" could be providing background information or context for the main action, which aligns with the use of PFV.CVB.

2. **PST.UNW as the main verb**: The main action of the sentence is "ik’in" (go-PST.UNW), which indicates the primary event. "Teqno" is not the main verb here, which makes it less likely to be PST.UNW.

3. **PFV.CVB in sequences of actions**: The sentence describes a sequence where the clever brother hears something and then goes behind them. "Teqno" could represent a completed action that leads to the subsequent action "ik’in" (go-PST.UNW), supporting the use of PFV.CVB.
\end{quote}

In the following case, the instruction-augmented system makes a mistake due to misrecognition of the main verb and labels `do' as a converb rather than a main verb.

\begin{exe}
    \ex
\gll Bo\textcrlambda{}\textcrlambda{}o zaman bik’inosi bara xediw bet’erbaqi boda ik’anix muk’ur oyn\\
III-middle time III-go-ANT.CVB wife-ERG husband life III-do-INF go-MSD-AD.ESS confess \textit{do-PST.UNW}\\
\glt `After some time has passed, the wife confessed that she went to live with her husband.'
\end{exe}

The chain of thought produced by ChatGPT labels `persuade', which is not inflected, as the main verb, and \textit{oyn} as a subordinate verb--- in fact, it acts as the head (auxiliary) verb. The issue, as discussed in the main text, is that ChatGPT tends to prioritize the English translation over the Tsez sentence--- the instructions state that "the PST.UNW verb typically follows the PFV.CVB verb if both are present", but also that the "main verb" should be tagged as PST.UNW. Here, the information about ordering (which would have given the right answer) is deemphasized in favor of the "main verb" criterion, which is incorrect due to the auxiliary. (The inferred tag sequence is shown in the excerpt. The retrieval system retrieves the primary meaning `persuade' rather than `confess' for \textit{muk'ur}, but this is corrected in the final gloss.) 

\begin{quote}
\footnotesize
Translation: 
1. **Bo\textcrlambda{}\textcrlambda{}o** - noun (III-middle)\\
2. **zaman** - noun (time)\\
3. **bik’inosi** - verb (I.PL-go-ANT.CVB)\\
4. **bara** - noun (wife-ERG)\\
5. **xediw** - noun (husband)\\
6. **bet’erbaqi** - noun (life)\\
7. **boda** - verb (III-do-INF)\\
8. **ik’anix** - verb (go-MSD-AD.ESS)\\
9. **muk’ur** - verb (persuade)\\
10. **oyn** - verb (?)\\

1. **PFV.CVB in subordinate clauses**: The candidate gloss suggests that "bik’inosi" (I.PL-go-ANT.CVB) is a subordinate clause providing background information. The main action seems to be the wife's confession about going to live with her husband.
 "Oyn" could potentially be a PFV.CVB if it provides context or background to the main action.

\ldots
\end{quote}

\end{document}